%% file: JFPDA-article.tex
\documentclass{jfpda2008}

\usepackage{color}
\usepackage{verbatim}

\usepackage{subfigure} 

\usepackage{ifthen}

\usepackage{natbib}

\usepackage[french]{babel}
\usepackage[utf8]{inputenc}
\usepackage[T1]{fontenc}

\def \jfpda {jfpda}

\usepackage{graphicx}
\graphicspath{{./Images/}}
\usepackage{moreverb}
\usepackage[normalem]{ulem}
\usepackage{amsmath,amsfonts}
\usepackage{amssymb}
\usepackage{bm}
\usepackage{mdwlist}

\usepackage[ruled,vlined,linesnumbered]{algorithm2e} 
\ifx \DontPrintSemicolon \undefined
 \def \DontPrintSemicolon {\dontprintsemicolon}
\fi
\newcommand{\acom}[1]{\tcc{\scriptsize #1}}
\usepackage{scalefnt}

\input{commands-fr}

\newcommand\shortcite[1]{\citet{#1}}

\newcommand{\pseudocode}[1]{}

\shorttitle{Les POMDP font de meilleurs hackers}

\shortouvrage{JFPDA 2012}

\def\mytitle{Les POMDP font de meilleurs hackers:\\ Tenir compte de
  l'incertitude dans les tests de pénétration}

\newcommand\confpaper[1]{}

\renewcommand{\cite}[1]{\citep{#1}}

\title{\mytitle}

\author{Carlos Sarraute\inst{1}, Olivier Buffet\inst{2}, J\"org Hoffmann\inst{3}}

\institute{
  Core Security \& ITBA\\
  Buenos Aires, Argentina\\
  \texttt{carlos@coresecurity.com}\\
  \texttt{http://corelabs.coresecurity.com/}
  \and
  INRIA / Université de Lorraine\\
  LORIA, Nancy, France\\
  \texttt{olivier.buffet@loria.fr}\\
  \texttt{http://www.loria.fr/$\sim$buffet/}
  \and
  Saarland University\\
  Saarbr\"ucken, Germany\\
  \texttt{hoffmann@cs.uni-saarland.de}\\
  \texttt{http://www.loria.fr/$\sim$hoffmanj/}
}

\begin{document}

\maketitle

\begin{abstract}
Les tests de pénétration sont une méthodologie pour évaluer la
sécurité d'un réseau en générant et exécutant de possibles attaques informatiques. 
Automatiser cette tâche permet de réaliser des tests réguliers et
systématiques. Une question clef est: ``Comment générer ces attaques
?'' Ce problème se formule naturellement comme de la planification
dans l'incertain, plus précisément avec une connaissance incomplète de
la configuration du réseau. Les travaux antérieurs emploient de la
planification classique, et requièrent de coûteux pré-traitements
réduisant cette incertitude par l'application extensive de méthodes de
scan. Au contraire, nous modélisons ici le problème de la
planification d'attaques à travers des processus de décision
markoviens partiellement observables (POMDP). Ceci nous permet de
raisonner à propos de la connaissance disponible, et d'employer
intelligemment les actions de scan comme faisant partie de
l'attaque. Comme on s'y attendrait, cette solution précise ne passe
pas à l'échelle. Nous concevons donc une méthode qui repose sur les
POMDP pour trouver de bonnes attaques sur des machines individuelles,
lesquelles sont recomposées en une attaque sur le réseau
complet. Cette décomposition exploite la structure du réseau dans la
mesure du possible, faisant des approximations ciblées (seulement) où
cela est nécessaire. En évaluant cette méthode sur un jeu de tests
industriels adaptés convenablement, nous démontrons son efficacité à
la fois en temps de calcul et en qualité de la solution.
\end{abstract}

\input{introduction-fr}

\input{preliminaries-fr}

\input{POMDP-model-fr}

\input{decomposition-algorithm-fr}

\input{experiments-fr}

\input{conclusion-fr}

\bibliography{planning}

\end{document}

%% file: commands-fr.tex
\newcommand{\commentout}[1]{}

\newcommand{\cf}{cf.}

\def\cS{\mathcal{S}}

\def\cO{\mathcal{O}}

\def\cA{\mathcal{A}}

\newcommand{\component}{\ensuremath{C}}
\newcommand{\rewardfn}{\ensuremath{r}}
\newcommand{\pivotingrewardfn}{\ensuremath{pr}}
\newcommand{\reward}{\ensuremath{R}}
\newcommand{\pivotingreward}{\ensuremath{pR}}
\newcommand{\pathreward}{\ensuremath{pathR}}
\newcommand{\start}{\ensuremath{*}}
\newcommand{\emptyfirewall}{\ensuremath{\emptyset}}

\ifx \theorem \undefined
\newtheorem{theorem}{Théorème}
\newtheorem{proposition}{Proposition}

\newtheorem{example}{Exemple}

{\hfill \rule[0.3ex]{1ex}{1ex} \par \addvspace{\bigskipamount}}

{\hfill \rule[0.3ex]{1ex}{1ex} \par \addvspace{\bigskipamount}}

\fi

\definecolor{Blue}{rgb}{0,0.16,0.90}
\definecolor{Red}{rgb}{0.90,0.16,0}
\definecolor{DarkBlue}{rgb}{0,0.08,0.45}
\definecolor{ChangedColor}{rgb}{0.9,0.08,0}
\definecolor{CommentColor}{rgb}{0.2,0.8,0.2}
\definecolor{ToDoColor}{rgb}{0.1,0.2,1}

\newcommand{\ournote}[1]{}
\newcommand{\todo}[1]{}
\newcommand{\joerg}[1]{}
\newcommand{\olivier}[1]{}
\newcommand{\carlos}[1]{}

%% file: introduction-fr.tex
\section{Introduction}
\label{introduction}

Les tests de pénétration (le {\em Penetration Testing}, ou {\em
  pentesting}) sont une méthodologie pour évaluer la sécurité d'un
réseau en générant et exécutant de possibles attaques exploitant des
vulnérabilités connues des systèmes d'exploitation et des applications
(voir, par exemple, \cite{ArcGra04}). Automatiser cette tâche permet
des tests réguliers et systématiques sans un travail humain prohibitif
et rend le pentesting plus accessible aux non-experts. Une question
clef est: ``Comment générer ces attaques ?''

Une façon naturelle de résoudre cette question est sous la forme d'un
problème de {\em planification d'attaques}. C'est connu dans la
communauté de la planification automatique comme le domaine de la
``Cyber Sécurité'' \cite{BodGohHaiHar05}. Indépendamment (mais bien
plus tard), l'approche a été mise en avant aussi par Core Security
\cite{LucSarRic10}, une compagnie de l'industrie du pentesting. Sous
cette forme, la planification d'attaques est très technique,
s'attachant aux détails de bas niveau de la configuration système qui
sont pertinents pour les vulnérabilités. Ici nous nous intéressons
exclusivement à ce cadre. Nous considérons le pentesting régulier
automatique tel que conduit dans l'outil ``Core Insight Enterprise''
de Core Security. Nous utiliserons le terme ``planification
d'attaques'' dans ce sens.

\shortcite{LucSarRic10} encodent la planification
d'attaques en PDDL et utilise des planificateurs prêt-à-l'emploi. Ceci
est, en soi, déjà utile -- en fait, c'est actuellement employé
dans le produit commercial Core Insight Enterprise, reposant sur une
variante de Metric-FF \cite{hoffmann:jair-03}. Toutefois, cette
approche est limitée par son incapacité à gérer l'incertitude. L'outil
de pentesting ne peut être à jour sur tous les détails de la
configuration de chaque machine dans le réseau, lesquelles sont
maintenues par des utilisateurs individuels.

Core Insight Enterprise résoud cela par l'utilisation extensive de
méthodes de {\em scan} comme pré-traitement à la planification, qui ne
considère alors que des {\em exploits}, c'est-à-dire des actions de
piratage modifiant l'état du système. Les inconvénients de cela sont
(a) ce pré-traitement entraine des coûts importants en termes de temps
d'exécution et de trafic réseau, et (b) de toute façon, puisque les
scans sont imparfaits, il reste une incertitude résiduelle (Metric-FF
est lancé sur la base de la configuration qui semble la plus
probable). Un travail antérieur \cite{SarRicLuc11} a résolu (b) en
associant à chaque exploit une probabilité de succès. Cela ne permet
pas de modéliser les dépendences entre exploits, et requiert quand
même des scans extensifs (pour obtenir des probabilités de succès
réalistes) et ne résoud donc pas (a). Ici, nous fournissons la
première solution capable de résoudre à la fois (a) et (b), mêlant
intelligemment scans et exploits comme un vrai hacker le
ferait. L'intuition fondamentale est que le pentesting peut être
modélisé naturellement comme la résolution d'un POMDP.

Nous encodons la connaissance incomplète comme une incertitude sur
l'état, modélisant ainsi les configurations possibles du réseau à
l'aide d'une distribution de probabilités. Scans et exploits sont
déterministes puisque leurs résultats dépendent seulement de l'état
dans lequel ils sont exécutés. Des récompenses négatives encodent le
coût (la durée) des scans et exploits; des récompenses positives
encodent la valeur des cibles atteintes. Ce modèle incorpore des
pare-feux, des effets de bord préjudiciables des exploits (plantage de
programmes ou de machines complètes), et les dépendances entre
exploits reposant sur des vulnérabilités similaires.

Les résolveurs de POMDP ne passent pas à l'échelle pour de grands
réseaux. Cela n'est pas surprenant -- même le modèle en entrée croît
exponentiellement avec le nombre de machines. Nous montrons comment
résoudre cela en exploitant la structure du réseau. Nous voyons les
réseaux comme des graphes dont les nœuds sont des sous-réseaux
complètement connectés, et dont les arcs encodent les connections
entre ceux-ci, filtrées par des pare-feux. Nous décomposons ce graphe
en composants bi-connectés. Nous approximons les attaques sur ces
composants en combinant des attaques sur des sous-réseaux
individuels. Nous approximons ces dernières en combinant des attaques
sur des machines individuelles. Ces approximations sont conservatives,
c'est-à-dire qu'elle ne surestiment jamais la valeur de la politique
retournée. Les attaques sur des machines individuelles sont modélisées
et résolues comme des POMDP, et les solutions sont propagées vers les
plus hauts niveaux. Nous évaluons cette approche sur un jeu de tests
industriels de Core Insight Enterprise, montrant que, en comparaison
avec un modèle POMDP global, elle améliore largement le temps de
calcul à un faible coût en termes de qualité d'attaque.

Nous présentons ci-après quelques préliminaires. Nous décrivons
ensuite notre modèle POMDP, notre algorithme de décomposition, et nos
résultats expérimentaux, avant de conclure cet article.

%% file: preliminaries-fr.tex
\section{Préliminaires}
\label{preliminaries}

Nous présentons quelques notions sur la structure du réseau et le
pentesting, et donnons un rapide aperçu des POMDP.

\subsection{Structure du réseau}

Les réseaux informatiques peuvent être vus comme des graphes orientés
dont les nœuds sont donnés par un ensemble $M$ de \emph{machines}, et
dont les arcs sont des connections entre paires de $m\in
M$. Toutefois, en pratique, ces graphes de réseaux ont une structure
particulière. Ils tendent à être constitués de \emph{sous-réseaux},
c'est-à-dire de groupes $N$ (pour {\em Network}) de machines dans
lesquels chaque $m\in N$ est directement connecté à chaque $m'\in
N$. A l'inverse, tout sous-réseau $N$ n'est pas connecté à n'importe
quel autre sous-réseau $N'$ et, typiquement, si une telle connexion
existe, elle est alors filtrée par un \emph{pare-feu}.

Du point de vue d'un attaquant, les pare-feux filtrent les connexions
et limitent ainsi les attaques qui peuvent être exécutées quand on
essaye de pénétrer un sous-réseau $N'$ depuis un autre sous-réseau
$N$. D'autre part, une fois que l'attaquant a réussi à s'introduire
dans un sous-réseau $N$, l'accès à toutes les machines dans $N$ est
facilité. Ainsi, une représentation naturelle du réseau, du point de
vue de la planification d'attaques, est celle d'un graphe dont les nœuds
sont des sous-réseaux, et dont les arcs sont annotés avec des
pare-feux $F$ (pour {\em Firewall}). Nous appellerons ici ce graphe le
\emph{réseau logique} $LN$ (pour {\em Logical Network}), et dénotons
ces arcs avec $N \xrightarrow{F} N'$.

Nous formalisons les pare-feux comme des ensembles de règles décrivant
quels types de communications (par exemple des ports) sont
interdits. Ainsi, de plus petits ensembles correspondent à des
pare-feux ``plus faibles'', et le \emph{pare-feu vide} ne bloque
aucune communication.

On remarque que, dans notre modèle POMDP, nous ne prévoyons pas
l'escalade de privilège ou l'obtention de mots de passe. Cela peut à
la place être modélisé au niveau de $LN$. Différents niveaux de
privilèges sur la même machine $m$ peuvent être encodés à l'aide de
différentes copies de $m$. Si le contrôle de $m$ permet la
récupération de mots de passe, alors $m$ peut être connectée à l'aide
de pare-feux vides aux machines $m'$ qui peuvent être atteintes en
utilisant ces mots de passe, ou plus précisément à des copies à
haut-privilèges de ces $m'$.

\subsection{Les tests de pénétration}

L'incertitude dans le pentesting apparaît parce qu'il est impossible
de garder trace de tous les détails de \emph{configuration} des
machines individuelles, c'est-à-dire des versions exactes des
programmes installés, etc. Toutefois, on peut sans risque faire
l'hypothèse que l'outil de pentesting connaît la structure du réseau,
c'est-à-dire le graphe $LN$ et le filtrage effectué par chaque
pare-feu: les changements à ce niveau sont peu fréquents et peuvent
être facilement enregistrés.

L'objectif du pentesting est de prendre contrôle de certaines machines
(au contenu critique) du réseau. A n'importe qu'elle moment, chaque
machine a un {\em statut} unique. Une machine \emph{contrôlée} $m$ a
déjà été piratée. Une machine \emph{atteinte} $m$ est connectée à une
machine contrôlée, c'est-à-dire que soit $m$ est dans un sous-réseau
$N$ dont une machine est contrôlée, soit $m$ est dans un sous-réseau
$N'$ avec un arc de $LN$, $N \xrightarrow{F} N'$, tel que l'une des
machines de $N$ est contrôlée. Toutes les autres machines sont
\emph{non atteintes}. L'algorithme commence avec une machine
contrôlée, dénotée ici par \start.\footnote{Par simplicité, nous
  dénoterons \start\ comme un nœud séparé de $LN$. Si \start\ fait
  partie du sous-réseau $N$, cela implique de transformer $N \setminus
  \{\start\}$ en un nœud séparé de $LN$, connecté à \start\ à l'aide
  d'un pare-feu vide.} Nous utiliserons la situation suivante (petite
mais réaliste) comme exemple dans la suite de cet article.

\begin{example}\label{exp:running-preliminaries}
  L'attaquant a déjà piraté une machine $m'$, et souhaite maintenant
  attaquer une machine $m$ dans le même sous-réseau. L'attaquant
  connaît deux exploits: \emph{SA}, l'exploit ``Symantec Rtvscan
  buffer overflow''; et \emph{CAU}, l'exploit ``CA Unicenter message
  queuing''. SA cible une version particulière de ``Symantec
  Antivirus'', lequel écoute usuellement le port 2967. CAU cible une
  version particulière de ``CA Unicenter'', qui écoute usuellement sur
  le port 6668. Les deux fonctionnent seulement si un mécanisme de
  protection appelé \emph{DEP} (``Data Execution Prevention'') est
  désactivé.
\end{example}

Si SA échoue, alors il est probable que CAU échouera aussi (parce que
DEP est activé). L'attaquant ferait alors mieux d'essayer autre
chose. Atteindre un tel comportement requiert que le plan d'attaque
observe les résultats des actions, et réagisse en conséquence. La
planification classique (qui suppose une connaissance parfaite du
monde au moment de la planification) ne peut accomplir cela.

En outre, les scans de ports -- des actions d'observation testant si
un port particulier est ouvert ou non -- devraient être utilisés
seulement si l'on a réellement l'intention d'exécuter un exploit en
rapport. Ici, si nous commençons avec SA, nous devrions scanner
seulement le port 2967. Nous pouvons obtenir un tel comportement à
travers l'utilisation de POMDP. En revanche, pour réduire
l'incertitude, la planification classique requiert un pré-traitement
exécutant \emph{tous} les scans possibles. Dans cet exemple, il n'y en
a que deux -- les ports 2967 et 6668 -- mais en général il y en a un
grand nombre, ce qui provoque un important trafic réseau et un
important temps d'attente.

\subsection{Les POMDP}

Les processus décisionnels de Markov partiellement observables (POMDP)
sont usuellement définis \cite{Monahan82,KaeLitCas-aij98} par un uplet
$\langle \cS, \cA, \cO, T, O, r, b_0 \rangle$. Si le système est dans
un état $s \in \cS$ (l'{\em espace d'état}), et l'agent effectue une
action $a\in \cA$ (l'{\em espace d'action}), alors en résulte (1) une
transition vers un état $s'$ selon la {\em fonction de transition}
$T(s,a,s')=Pr(s'|s,a)$, (2) une observation $o\in\cO$ (l'{\em espace
  d'observation}) selon la {\em fonction d'observation}
$O(s',a,o)=Pr(o|s',a)$ et (3) une {\em récompense} scalaire
$r(s,a,s')$. $b_0$, la {\em croyance initiale}, est une distribution
de probabilité sur $\cS$.

L'agent doit trouver une {\em politique} de décision $\pi$
choisissant, à chaque étape, la meilleure action en fonction de ses
observations passées et des actions de manière à maximiser ses gains
futurs, lesquels nous mesurons ici à travers le total des récompenses
accumulées. La valeur espérée d'une politique optimale est dénotée
$V^*$.

L'agent raisonne typiquement sur l'état caché du système en utilisant
un {\em état de croyance} $b$, une distribution de probabilité sur
$\cS$. Pour nos expérimentations nous utilisons SARSOP
\cite{KurHsuLee08}, un algorithme à base de points de l'état de l'art,
c'est-à-dire un algorithme approchant la fonction de valeur comme
l'enveloppe supérieure d'un ensemble d'hyperplans, lesquels
correspondent à une sélection d'états de croyance particuliers
(appelés ``points'').

%% file: POMDP-model-fr.tex
\section{Modèle POMDP}
\label{POMDP-model}

Une version préliminaire de notre modèle POMDP a été présentée dans le
workshop SecArt'11 \cite{SarBufHof11}. Le lecteur peut se référer à
cet article pour un exemple plus détaillé listant des modèles de
transition et d'observation complets pour certaines actions, et
illustrant l'évolution des états de croyance quand ces actions sont
appliquées. Dans ce qui suit, nous gardons des descriptions brèves par
soucis de gain d'espace.

\subsection{\'{E}tats}
\label{POMDP-model:states}

Plusieurs aspects du problème -- notamment la structure du réseau et
les règles de filtrage des pare-feux -- sont connus et statiques. Les
variables encodant ces aspects peuvent être éliminées lors d'un
pré-traitement, et ne sont pas incluses dans notre modèle.

Les états saisissent le statut de chaque machine
(contrôlée/atteinte/non atteinte). Pour les machines non contrôlées,
ils spécifient aussi la configuration logicielle (système
d'exploitation, serveurs, ports ouverts, \dots). Nous spécifions les
programmes vulnérables, ainsi que les programmes qui peuvent fournir
de l'information à propos de ceux-ci (par exemple, le mécanisme de
protection ``DEP'' de notre exemple est pertinent pour les deux
exploits). Les états indiquent aussi si une machine ou un programme
donné a planté.

Finalement, nous introduisons un état spécial, \emph{terminal}, dans le
modèle POMDP (de l'ensemble du réseau, pas des machines
individuelles). Cet état correspond à l'abandon de l'attaque, quand,
pour chaque action disponible (s'il en reste), le bénéfice potentiel
ne compense pas les coûts d'action.

\begin{example}\label{exp:running-states}
  Les états décrivent la machine attaquée $m$. Pour simplifier, nous
  ferons l'hypothèse que les exploits ne risquent pas ici de planter
  la machine (voir aussi la sous-section suivante). A part l'état
  terminal et l'état représentant le fait que $m$ est contrôlée, les
  états spécifient quels programmes (``SA'' ou ``CAU'') sont présents,
  s'ils sont vulnérables, et si ``DEP'' est activé. Chaque application
  écoute sur un port différent, donc un port est ouvert ssi
  l'application associée est présente, et nous n'avons pas besoin de
  modéliser les ports séparément. Nous avons ainsi un total de $20$ états:\\ {
\begin{minipage}[t]{.24\linewidth}
\begin{verbatim} 
1 terminal
2 m_controlled
\end{verbatim}
\end{minipage}
\begin{minipage}[t]{.34\linewidth}
\begin{verbatim}
 3 m_none
 4 m_CAU
 5 m_CAU_Vul
 6 m_SA
 7 m_SA_CAU
 8 m_SA_CAU_Vul
 9 m_SA_Vul
10 m_SA_Vul_CAU
11 m_SA_Vul_CAU_Vul
\end{verbatim}
\end{minipage}
\begin{minipage}[t]{.30\linewidth}
\begin{verbatim}
12 m_DEP_none
13 m_DEP_CAU
14 m_DEP_CAU_Vul
15 m_DEP_SA
16 m_DEP_SA_CAU
17 m_DEP_SA_CAU_Vul
18 m_DEP_SA_Vul
19 m_DEP_SA_Vul_CAU
20 m_DEP_SA_Vul_CAU_Vul
\end{verbatim}
\end{minipage}
}
\end{example}

En résumé, les états de chaque machine $m$ sont essentiellement des
uplets de valeurs de statuts pour chaque programme pertinent. Les
états du système global sont alors des uplets des ces états de
machines, avec une entrée pour chaque $m \in M$. L'espace d'état
énumère ces uplets. Dit autrement, l'espace d'état est factorisé d'une
façon naturelle, par programmes et par machines. Une option évidente
est donc de modéliser et résoudre le problème en utilisant des POMDP
factorisés \cite{HanFen00}. Nous n'avons pas encore essayé cela; notre
générateur de modèle POMDP énumère les états en interne et fournit ce
modèle à SARSOP.\footnote{Notons que cette approche permet certaines
  optimisations non-triviales: certains des états dans
  l'exemple~\ref{exp:running-states} pourraient être fusionnés. Si DEP
  est activé, alors il importe peu que CAU/SA soient vulnérables. Par
  souci de brièveté, nous ne discutons pas ici ce point en détail.}

La nature factorisée de notre problème implique aussi que l'espace
d'état est énorme. Dans un cadre réaliste, l'ensemble $C$ des uplets
de configurations possibles pour chaque machine $m\in M$ est très
grand, donnant un espace d'état énorme $|\cS|=O(|C|^{|M|})$. En
pratique, nous n'exécuterons des POMDP que sur des machines seules,
c'est-à-dire pour $|M|=1$.

\subsection{Actions}
\label{POMDP-model:actions}

Pour atteindre l'état terminal, nous avons besoin d'une action
\emph{terminate} (terminer) indiquant que l'on abandonne l'attaque.

Il y a deux principaux types d'actions, les \emph{scans} et les
\emph{exploits}, les deux devant être ciblées vers des machines
atteignables.
Les scans peuvent être des actions de détection d'OS ou des scans de
ports. Dans la plupart des cas, ils n'auront pas d'effet sur l'état de
la machine cible. Leur objectif est d'acquérir de la connaissance à
propos de la configuration de la machine à l'aide d'une observation
qui permet typiquement d'éliminer certains états de la croyance (par
exemple, observer que l'OS doit être une certaine version de Windows
XP).
Les exploits font usage d'une vulnérabilité -- si elle est présente --
pour prendre le contrôle d'une machine. Le résultat de cet exploit est
observé par l'attaquant, de sorte que l'échec d'un exploit peut, comme
un scan, donner de l'information à propos de la configuration (par
exemple, que le mécanisme de protection est probablement actif). Pour
une minorité d'exploits, une tentative ratée plante la machine.

Pour toutes les actions, le résultat est déterministe: quelle
observation est retournée, et si un exploit réussit/échoue/plante, est
déterminé de manière unique par la configuration de la machine cible.

\begin{example}\label{exp:running-actions}
Dans notre exemple, il y a cinq actions possibles:\\ {
\begin{minipage}[t]{.4\linewidth}
\begin{verbatim}
m_exploit_SA
m_exploit_CAU
m_scan_port_2967
m_scan_port_6668
terminate
\end{verbatim}
\end{minipage}
}

\smallskip
\noindent
Le modèle POMDP spécifie, pour chaque état dans
l'exemple~\ref{exp:running-states}, le résultat de chaque action. Par
exemple, {\verb+m_exploit_SA+} réussit si et seulement si SA est
présent et vulnérable, et DEP est désactivé. Ainsi, quand appliquée à
l'état 9, 10 ou 11, {\verb+m_exploit_SA+} résulte en l'état 2, et
retourne l'observation {\verb+succeeded+}. Appliquée à n'importe quel
autre état, {\verb+m_exploit_SA+} laisse l'état inchangé, et
l'observation est {\verb+failed+}.
\end{example}

Les résultats des actions dépendent aussi du pare-feu (s'il y en a un)
érigé entre le pentesteur et la cible. Si le pare-feu filtre le port
associé à une action, alors celle-ci est inutilisable: son modèle de
transition laisse l'état inchangé, et aucune observation n'est
retournée. Par exemple, si le pare-feu $F$ filtre le port 2967, alors
{\verb+m_scan_port_2967+} et {\verb+m_exploit_SA+} sont inutilisables
à travers $F$, mais peuvent être employées dès qu'une machine derrière
$F$ est sous contrôle.

\subsection{Récompenses}
\label{POMDP-model:rewards}

Aucune récompense n'est obtenue quand on utilise l'action
\emph{terminate} ou dans l'état terminal.

La récompense instantanée pour n'importe quelle action scan/exploit
dépend de la transition qu'elle entraîne depuis l'état courant. Notre
modèle simple est de décomposer additivement la récompense instantanée
$\rewardfn(s,a,s')$ en $\rewardfn(s,a,s') = r_e(s,a,s') + r_t(a) +
r_d(a)$. Ici, (i) $r_e(s,a,s')$ est la valeur de la machine attaquée
dans le cas où la transition $(s,a,s')$ correspond à un exploit
réussi, et vaut $0$ pour toute autre transition; (ii) $r_t(a)$ est un
coût qui dépend de la durée de l'action; et
(iii) $r_d(a)$ est un coût qui reflète le risque de détection quand on
utilise cette action. (iii) est orthogonal au risque de planter un
programme ou une machine, ce que, comme décrit, nous modélisons comme
un possible résultat d'exploits. Notons que (ii) et (iii) peuvent être
corrélés; toutefois, il n'y a pas bijection entre durée et risque de
détection d'un exploit, il est donc pertinent d'être capable de
distinguer les deux. Finalement, notons que (i) résulte de la somme
des récompenses pour les exploits réussis sur différentes machines. Ce n'est
pas une hypothèse contraignante: on peut récompenser le fait de
pénétrer dans \emph{[$m_1$ OU $m_2$]} en introduisant une nouvelle
machine virtuelle, accessible sans coût depuis aussi bien $m_1$ que
$m_2$.

\begin{example}\label{exp:running-rewards}
  Dans notre exemple, nous fixons $r_e=1000$ en cas de succès, $0$
  sinon; $r_t=-10$ pour toutes les actions; and $r_d=0$ (aucun risque
  de détection). Nous verrons ci-dessous quel effet ces réglages ont
  sur une politique optimale.
\end{example}

Puisque toutes les actions sont déterministes, il n'est pas utile de
les répéter sur la même cible à travers le même pare-feu -- cela ne
produira pas de nouveaux effets ou n'apportera pas de nouvelles
informations. En particulier, les récompenses positives ne peuvent
être reçues plusieurs fois. Ainsi, des comportements cycliques
encourent des coûts négatifs infinis. Cela implique que le retour
espéré d'une politique optimale est fini même sans facteur
d'atténuation.\footnote{En fait, le problème tombe dans la classe des
  \emph{problèmes de chemin stochastique le plus court}
  \cite{BerTsi96}.}

\subsection{Concevoir la croyance initiale}
\label{POMDP-model:initial-belief}

Un test de pénétration est effectué à intervalles de temps
réguliers. La croyance initiale -- notre connaissance du réseau quand
on commence le pentest -- dépend de (a) ce qui était connu à la fin du
pentest précédent, et de (b) ce qui peut avoir changé depuis. Nous
faisons l'hypothèse pour des raisons de simplicité que la connaissance
(a) est parfaite, c'est-à-dire qu'une configuration concrète $I(m)$
est affectée à chaque machine $m$ au temps $0$ (du dernier pentest). Nous
calculons alors la croyance initiale comme une fonction $b_0(I,T)$ où
$T$ est le nombre de jours passés depuis le dernier
pentest. L'incertitude dans cette croyance vient de ne pas savoir
quelles mises à jour logicielles ont été appliquées. Nous faisons
l'hypothèse que les mises à jour sont effectuées indépendamment sur
chaque machine (hypothèse simplificatrice, mais raisonnable étant
donné que les mises à jour sont contrôlées par des utilisateurs
individuels).

Un modèle simple de mises à jour \cite{SarBufHof11} encode l'évolution
incertaine de chaque programme indépendamment au moyen d'une chaîne de
Markov. Les états de chaque chaîne correspondent aux différentes
versions du programme, et les transitions modélisent les mises à jour
possibles du programme (avec des probabilités estimées que ces mises à
jour auront lieu). La croyance initiale est alors la distribution
résultant de cette chaîne après $T$ pas de temps.

\vspace{-0.05cm}
\begin{example}\label{exp:generating-initial-belief}

  \begin{figure}[t]

    \begin{minipage}{0.3\linewidth}
      \centerline{\includegraphics[width=0.9\linewidth]{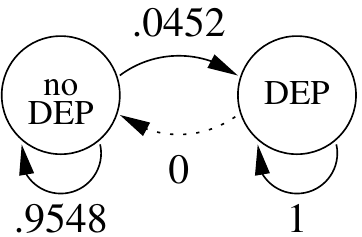}}
    \end{minipage}
    \hfill
    \begin{minipage}{0.3\linewidth}
      \centerline{\includegraphics[width=0.9\linewidth]{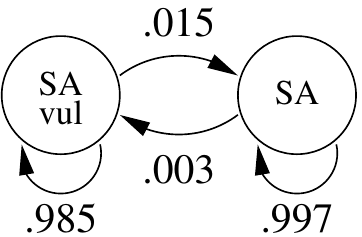}}
    \end{minipage}
%
\hfill
      \begin{minipage}{0.3\linewidth}
        \centerline{\includegraphics[width=0.9\linewidth]{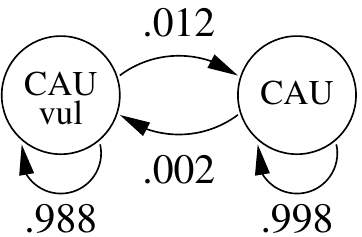}}
      \end{minipage}
      \caption{Les trois chaînes de Markov indépendantes utilisées
        pour modéliser le mécanisme de mise à jour de notre réseau
        exemple.}
    \label{fig:MarkovChains}
  \end{figure}

  Dans notre exemple, les trois composants de l'unique machine sont
  DEP, CAU et SA. Ils sont mis à jour à l'aide de trois chaînes de
  Markov indépendantes, chacune avec deux états, comme illustré sur la
  figure~\ref{fig:MarkovChains}. Les probabilités indiquent quelles
  sont les chances que la machine passe d'un état à un autre pendant
  un jour. Disons que l'on fixe $T=30$, et lancions les chaînes de
  Markov sur la configuration $I$ dans laquelle $m$ a le DEP
  désactivé, et SA et CAU sont tous deux vulnérables aux exploits de
  l'attaquant. Dans la croyance initiale résultante $b_0(I,T)$, DEP a
  des chances d'être activé; le poids des états 12--20 dans
  l'exemple~\ref{exp:running-states} est élevé dans $b_0$ ($>70\%$).
\end{example}
\vspace{-0.05cm}

Ici, nous utilisons ce modèle simple comme bloc de construction
élémentaire dans une méthode prenant en compte que la version $x$ du
programme A peut nécessiter la version $y$ ou $z$ du programme B. Nous
faisons l'hypothèse que les programmes sont organisés d'une manière
hiérarchique, le système d'exploitation étant à la racine d'un graphe
orienté acyclique, et un programme ayant comme parents les programmes
dont il dépend directement. Cela donne un réseau bayésien dynamique,
dans lequel chaque distribution de probabilité conditionnelle est
dérivé d'une chaîne de Markov $Pr(X_t=x'|X_{t-1}=x)$ %
filtrée par une fonction de compatibilité $\delta(X=x,parent_1(X)=x_1,
\dots, parent_k(X)=x_k)$, qui retourne $1$ ssi la valeur de $X$ est
compatible avec les versions des parents, $0$ sinon. Ce modèle de
mises à jour est raisonnable, mais évidemment pas encore réaliste; les
travaux futurs devront étudier de tels modèles en détail.

Nous illustrons maintenant comment le fait de raisonner avec les probabilités de
la croyance initiale permet d'obtenir le comportement intelligent désiré.

\vspace{-0.05cm}
\begin{example}\label{exp:running-initial-belief-policy}
  Supposons que nous calculons la croyance initiale $b_0(I,T)$ comme
  dans l'exemple~\ref{exp:generating-initial-belief}. Puisque le poids
  des états 12--20 est élevé dans $b_0$, si {\verb+m_exploit_SA+}
  échoue, alors la probabilité de succès de {\verb+m_exploit_CAU+} est
  réduite au point de ne plus valoir d'efforts, et que l'attaquant
  (une politique optimale) abandonne, c'est-à-dire qu'elle essaye une
  attaque différente non entravée par le DEP. Plus précisément,
  considérons $Pr($CAU$^+|$2967$^+)$, c'est-à-dire la probabilité que
  {\verb+m_exploit_CAU+} réussisse, après avoir observé que le port
  2967 est ouvert. Cela correspond au poids (A) des états 8 et 11 dans
  l'exemple~\ref{exp:running-states}, par rapport aux états (B) 6--11
  plus 15--20. Ce poids (A/B) est à peu près $20\%$. Ainsi la valeur
  espérée de {\verb+m_exploit_CAU+} dans cette situation est à peu
  près $100*0,2$ [récompense de succès] $- 10$ [coût d'action] $= 10$,
  cf.\ exemple~\ref{exp:running-rewards}, de sorte que l'action en
  vaut la peine. A l'inverse, supposons que {\verb+m_exploit_SA+} a
  déjà été essayée et a échoué. Alors (A) est réduit à l'état 8 seul,
  alors que (B) contient toujours (en particulier) tous les états DEP
  15--20. Ces derniers états ont un poids important, et donc
  $Pr($CAU$^+|$2967$^+,$SA$^-)$ n'est que d'environ $5\%$. Compte tenu
  de cela, la valeur espérée de {\verb+m_exploit_CAU+} est négative,
  et il est préférable d'appliquer {\em terminate} à la place.
\end{example}
\vspace{-0.05cm}

%% file: decomposition-algorithm-fr.tex
\section{L'algorithme de décomposition 4AL}
\label{decomposition-algorithm}

\ifthenelse{\isundefined{\jfpda}}{
\begin{figure*}[t]
\begin{tabular}{ccc}
    \scalebox{0.7}{
      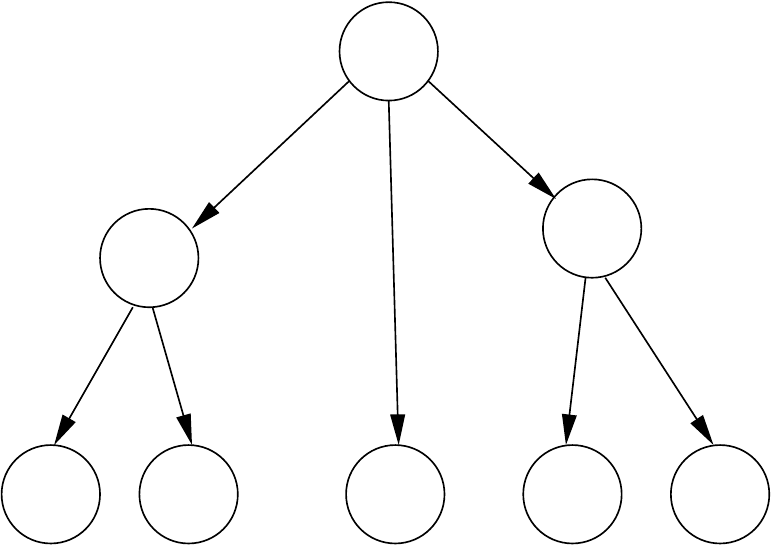
    } &
    \scalebox{0.7}{
      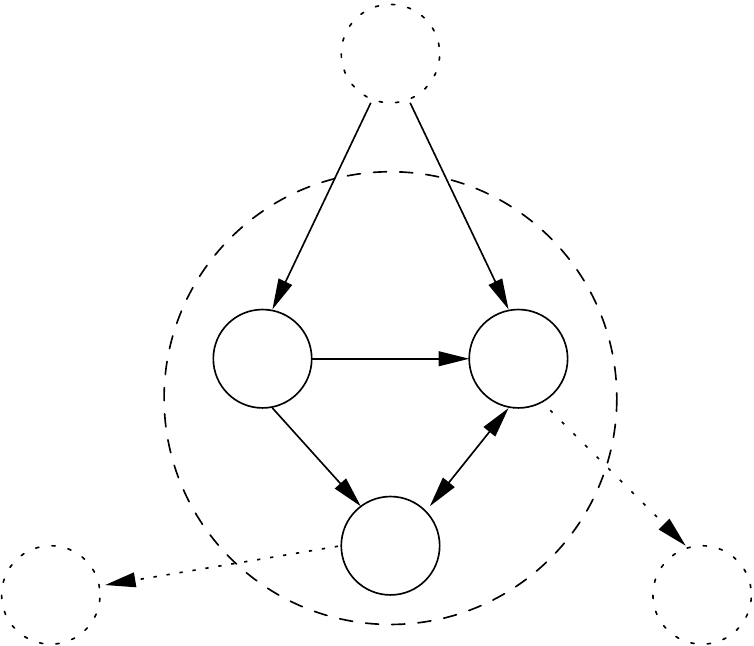
    } &
    \scalebox{0.7}{
      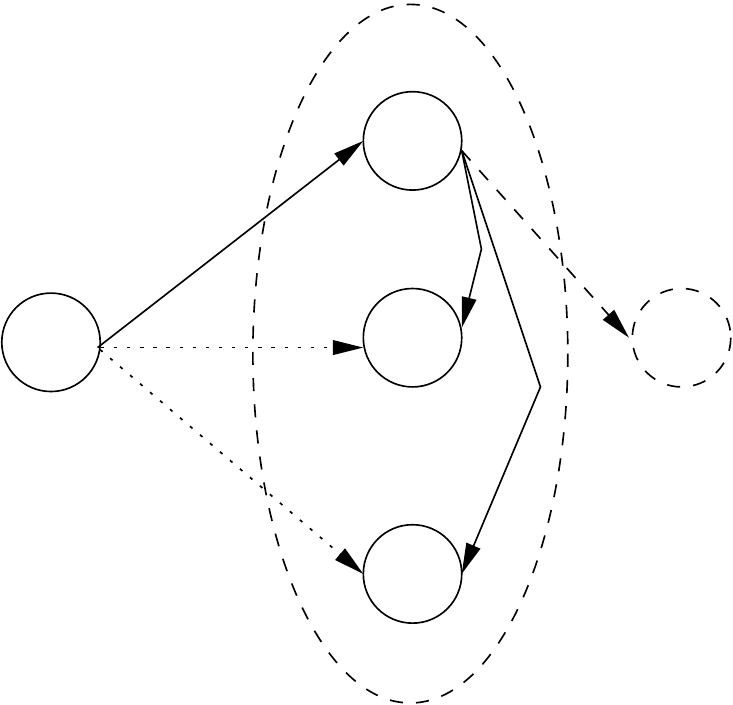
    } \\[0.1cm]
    (a) $LN$ arbre de composants $C$. & 
    (b) Chemins pour attaquer $C_1$. & 
    (c) Attaque de $N_3$ depuis $N_1$, en utilisant d'abord $m$.
  \end{tabular}
  \vspace{-0.35cm}
  \caption{Illustration des niveaux 1, 2 et 3 (de gauche à droite) de
    l'algorithme 4AL.}
  \label{fig:4ALlevels123}
  \vspace{-0.5cm}
\end{figure*}
}{
  \begin{figure}[t]
    \begin{minipage}{0.45\linewidth}
      \centering
      \scalebox{0.7}{
        \input{Images/4AL1.pdf_t}
      }
      \\
      (a) $LN$ arbre de composants $C$.
    \end{minipage}
    \hfill  
    \begin{minipage}{0.45\linewidth}
      \centering
      \scalebox{0.7}{
        \input{Images/4AL2.pdf_t}
      }
      \\
      (b) Chemins pour attaquer $C_1$.
    \end{minipage}

    \medskip
    
    \begin{minipage}{1\linewidth}
      \centering
      \scalebox{0.7}{
        \input{Images/4AL3.pdf_t}
      }
      \\
      (c) Attaque de $N_3$ depuis $N_1$, en utilisant d'abord $m$.
    \end{minipage}
    \caption{Illustration des niveaux 1, 2 et 3 (de gauche à droite)
      de l'algorithme 4AL.}
  \label{fig:4ALlevels123}
\end{figure}
}

\begin{figure*}[t]
\centering
\begin{tabular}{cc}
\input{Algorithms/4AL1}
&
\input{Algorithms/4AL2}
\\
\input{Algorithms/4AL3}
&
\input{Algorithms/4AL4}
\end{tabular}
\label{fig:4ALcode}
\end{figure*}

Comme évoqué, les POMDP ne passent pas à l'échelle pour de grands
réseaux (cf.\ les expérimentations dans la section suivante). Nous
présentons maintenant une approche utilisant décomposition et
approximation pour surmonter ce problème, reposant sur les POMDP
seulement pour attaquer des machines individuelles. L'approche est
appelée \emph{4AL} parce qu'elle s'attache à l'attaque d'un réseau à 4
niveaux d'abstraction différents ({\em 4 abstraction levels}). 4AL est
un résolveur de POMDP spécialisé dans la planification d'attaques
telle que considérée ici. Son entrée est un réseau logique $LN$ et des
modèles POMDP encodant les attaques de machines individuelles. Sa
sortie est une politique (une attaque) pour le POMDP global encodant
$LN$, de même qu'une approximation de la fonction de valeur
globale. Nous donnons dans la suite un aperçu de l'algorithme, puis
précisons les détails techniques. Pour simplifier la présentation,
nous nous concentrerons sur l'approximation de la fonction de valeur,
et décrirons seulement brièvement comment construire la politique.

\subsection{Aperçu de 4AL et propriétés élémentaires}
\label{decomposition-algorithm:overview}

Les quatre niveaux de 4AL sont: (1) \emph{décomposer le réseau}, (2)
\emph{attaquer les composants}, (3) \emph{attaquer les sous-réseaux},
et (4) \emph{attaquer les machines individuelles}. Nous décrivons les
niveaux tour à tour avant de fournir des détails techniques. La
figure~\ref{fig:4ALlevels123} fournit des illustrations.

\begin{itemize}
\item \textbf{Niveau~1:} Décompose le réseau logique $LN$ en un arbre
  de composants bi-connectés, enraciné en \start. Dans l'ordre
  topologique inverse, appeler le niveau~2 sur chaque composant;
  propager les résultats vers le haut dans l'abre.
\end{itemize}

Chaque graphe se décompose en un unique arbre de composants
bi-connectés \cite{HopTar-cACM73}. Un composant bi-connecté est un
sous-graphe qui reste connecté quand on retire n'importe quel
arc. Dans le pentesting, cela signifie intuitivement qu'il y a plus
qu'une façon (plus d'un chemin) pour attaquer les sous-réseaux de ce
composant, ce qui nécessite de raisonner à propos du composant comme
un tout (ce qui est le travail du niveau~2). Au contraire, si retirer
le sous-réseau $X$ (par exemple, $N_2$ dans la
figure~\ref{fig:4ALlevels123} (b)) fait que le graphe est coupé en
deux morceaux ($C_2$ vs.\ le reste de $LN$, voir aussi la
figure~\ref{fig:4ALlevels123} (a)), alors \emph{toutes} les attaques
de \start\ vers l'un de ces sous-graphes (ici $C_2$) doivent d'abord
traverser $X$ (ici $N_2$). Ainsi la valeur espérée totale de l'attaque
peut être calculée par (1) le calcul la valeur d'une attaque de ce
sous-graphe ($C_2$) seul, et (2) l'ajout de ce résultat comme une
\emph{récompense de pivotement} pour récompenser la prise de contrôle
de $X$ ($N_2$). Dit autrement, nous ``propageons les résultats vers le
haut'' dans l'arbre représenté dans la figure~\ref{fig:4ALlevels123}
(a).

Il est important de noter que cette décomposition arborescente
résultera typiquement en une énorme réduction de complexité. Les
composants bi-connectés dans $LN$ ne proviennent que de clusters de
plus de $2$ sous-réseaux partageant une même machine pare-feu
(physique). De tels clusters tendent à être petits. Dans le scénario
de test réaliste utilisé par Core Security et ici dans nos
expérimentations, il n'y a qu'un seul cluster, de taille $3$. S'il n'y
a pas de cluster du tout, $LN$ est un arbre et le niveau~2 de 4AL
devient complètement trivial.

\begin{itemize}
\item \textbf{Niveau~2:} Etant donné le composant $\component$,
  considérons, pour chaque sous-réseau récompensé $N \in \component$,
  tous les chemins $P$ dans $\component$ qui atteignent $N$. En remontant
  chaque $P$ en arrière, appeler le niveau~3 sur chaque sous-réseau
  avec chaque pare-feu associé. Choisir le meilleur chemin pour chaque
  $N$. Agréger les valeurs de ces chemins sur tout $N$, en
  additionnant mais en ignorant les récompenses qui ont déjà été
  prises en compte par un chemin précédent dans la somme.
\end{itemize}

Dans le cas où un composant bi-connecté $\component$ contient plus
d'un sous-réseau, pour obtenir la meilleure attaque de $\component$,
en général nous n'avons d'autre choix que d'encoder le composant
entier comme un POMDP. Puisque ce n'est pas faisable, le niveau~2
considère des ``chemins d'attaque'' individuels à l'intérieur de
$\component$. N'importe quel chemin $P$ est équivalent à une séquence
d'attaques sur des sous-réseaux individuels; ces attaques sont
évaluées en utilisant le niveau~3. Nous considérons les nœuds
récompensés $N$ séparément, en énumérant les chemins d'attaque et en
choisissant le meilleur. Les valeurs des meilleurs chemins sont
agrégées sur tout $N$ d'une manière conservative (pessimiste), en
tenant compte de chaque récompense au plus une fois. Une
sous-estimation stricte a lieu dans le cas où les meilleurs chemins
pour certains nœuds récompensés ne sont pas disjoints: alors ces
attaques partagent une partie de leur coût, de sorte qu'une attaque
combinée a une récompense espérée plus élevée que la somme des
attaques indépendantes.

Dans la figure~\ref{fig:4ALlevels123} (b), $N_2$ et $N_3$ ont une
récompense de pivotement parce qu'ils permettent d'atteindre les
composants $C_2$ et $C_3$ respectivement. Si les meilleurs chemins
pour à la fois $N_2$ et $N_3$ passent par $N_1$ (parce que le pare-feu
$F^*_3$ est très strict), alors ces chemins ne sont pas disjoints,
dupliquant l'effort pour pénétrer dans $N_1$.

Évidemment, énumérer les chemins d'attaque dans $\component$ est
exponentiel en la taille de $\component$. C'est le seul point dans 4AL
-- à part évidemment les appels au résolveur de POMDP -- qui a un
temps d'exécution exponentiel dans le pire cas. En pratique, les
composants bi-connectés sont typiquement petits, \cf\ ci-dessus.

\begin{itemize}
\item \textbf{Niveau~3:} Etant donné un sous-réseau $N$ et un pare-feu
  $F$ à travers lequel attaquer $N$, pour chaque machine $m\in N$,
  approximer la récompense obtenue en attaquant $m$ d'abord. Pour
  cela, modifier la récompense de $m$ pour tenir compte du fait que,
  après avoir pénétré $m$, nous serons derrière $F$: appeler le
  niveau~4 pour obtenir les valeurs de tous les $m' \neq m$ avec un
  pare-feu vide; ajouter ensuite ces valeurs, plus toute récompense de
  pivotement, à la récompense de $m$ et appeler le niveau~4 sur ce $m$
  modifié avec le pare-feu $F$. Maximiser la valeur résultante sur
  tous les $m\in N$.
\end{itemize}

Considérons la figure~\ref{fig:4ALlevels123} (c). Quand on attaque $N$
(ici, $N_3$) depuis une machine derrière $F$ (ici, $F^1_3$), nous
avons à choisir quelle machine dans $N$ attaquer. Etant donné qu'on
s'engage sur un tel choix $m$, le problème de l'attaque devient celui
de pénétrer dans $m$ et après cela d'exploiter la connexion directe
vers n'importe quel $m \neq m' \in N$, et n'importe quel réseau
descendant (ici, $C_3$) vers lequel on peut pivoter. Comme décrit, on
peut traiter cela en combinant des attaques sur des machines
individuelles avec des récompenses modifiées. (La récompense de
pivotement pour les réseaux descendants est calculée au préalable par
les niveaux 1 et 2.)

Comme le niveau~2, le niveau~3 fait des approximations
conservatives. Il fixe un choix du $m\in N$ à attaquer. Au contraire,
la meilleure stratégie peut consister à basculer entre différents
$m\in N$ selon le succès de l'attaque jusqu'ici. Par exemple, si un
exploit a de grandes chances de réussir, alors il peut être payant de
l'essayer d'abord sur tous les $m$, avant d'essayer quoi que ce soit
d'autre.

\begin{itemize}
\item \textbf{Niveau~4:} Etant donnée une machine $m$ et un pare-feu
  $F$, modéliser le problème de planification d'attaques pour une
  seule machine comme un POMDP, et lancer un résolveur
  prêt-à-l'emploi. Mettre en cacher les résultats obtenus pour éviter
  les doubles emplois.
\end{itemize}

Cette dernière étape ne devrait pas avoir besoin d'explications. Le
modèle POMDP est créé comme décrit précédemment. Notons que le
niveau~3 peut, durant l'exécution de 4AL, appeler la même machine avec
le même pare-feu plus d'une fois. Par exemple, dans la
figure~\ref{fig:4ALlevels123} (c), quand on bascule vers l'attaque de
$m'_1$ au lieu de $m$, l'appel du niveau~4 avec $m'_k$ et un pare-feu
vide est répété.

Pour résumer, 4AL a un temps de calcul bas polynomial sauf pour
l'énumération de chemins dans les composants bi-connectés (niveau~2),
et résoudre des POMDP pour une seule machine (niveau~4). La
décomposition au niveau~1 ne subit aucune perte d'information. Les
niveaux 2 et 3 font des approximations conservatives, de sorte que, si
les solutions des POMDP sont conservatives (par exemple, optimales),
alors le résultat d'ensemble de 4AL est aussi conservatif.

\subsection{Détails techniques}
\label{decomposition-algorithm:technicalities}

Pour fournir une compréhension plus détaillée de 4AL, nous discutons
maintenant du pseudo-code de l'algorithme, fourni dans la
figure~\ref{fig:4ALcode}. Considérons d'abord
l'algorithme~\ref{alg:level1}. Il devrait être clair comment la
structure d'ensemble de l'algorithme correspond à notre discussion
précédente. Il appelle l'algorithme à temps linéaire de
\shortcite{HopTar-cACM73} (par la suite, HT) pour trouver la
décomposition. La boucle $i=k,\dots,1$ traite les composants dans
l'ordre topologique inverse. La fonction de récompense de pivotement
$\pivotingrewardfn$ encode la propagation de récompenses vers le haut
de l'arbre; cela ne devrait pas avoir besoin d'explications, à part
pour l'expression ``le parent'' de $C_i$ dans $LN$. Cette dernière
repose sur le fait que, après nettoyage (``clean-up'', ligne 2),
chaque composant a exactement un tel parent.

Pour expliquer le nettoyage, notons d'abord que HT travaille sur des
graphes non-orientés; quand nous l'appliquons, nous ignorons la
direction des arcs dans $LN$. Le résultat est un arbre non-orienté de
composants bi-connectés, où les \emph{nœuds-coupes} -- les nœuds dont
la suppression coupe le graphe en morceaux -- sont partagés entre
plusieurs composants. Dans la figure~\ref{fig:4ALlevels123} (b), par
exemple, $N_2$, avant nettoyage, appartient à la fois à $C_1$ et
$C_2$. Le nettoyage fixe la racine de l'arbre à \start, et affecte
chaque nœud-coupe au composant le plus proche de \start\ (par exemple,
$N_2$ est affecté à $C_1$); \start\ lui-même est transformé en un
composant séparé. En ré-introduisant l'orientation des arcs dans $LN$,
nous éliminons alors les nœuds non accessibles depuis \start. Ensuite,
nous retirons les arcs qui ne peuvent pas participer à des chemins
d'attaque non redondants partant de \start. Puisqu'aller \emph{vers}
\start\ dans la décomposition arborescente ramène nécessairement toute
attaque vers un nœud qu'elle a déjà visité (pénétré), après une telle
suppression les arcs entre les composants forment un arbre orienté
comme sur la figure~\ref{fig:4ALlevels123} (a). Chaque composant
non-racine $\component_i$ (par exemple $C_3$) a exactement un parent
composant $\component$ dans l'arbre nettoyé (par exemple $C_1$). Le
sous-réseau respectif $N \in \component$ (par exemple $N_3$) est un
nœud-coupe de $LN$. Ainsi, comme affirmé plus haut, $N$ est le
\emph{seul} nœud, dans $LN$, connecté à $\component_i$.

De manière évidente, toutes les attaques sur $\component_i$ doivent
passer à travers leur parent $N$. De plus, les nœuds et arcs supprimés
par nettoyage ne peuvent pas faire partie d'une attaque
optimale. Ainsi le niveau~1 est sans perte. Pour exprimer cela -- et
les autres propriétés de 4AL -- formellement, nous avons besoin de
quelques notations. Nous utiliserons $V^*$ pour désigner la valeur
espérée réelle (optimale) d'une attaque, et $V$ pour désigner
l'approximation 4AL. L'objet attaqué est donné en argument. Par
exemple, $V^*(LN)$ est la valeur espérée pour attaquer $LN$;
$V(C,\pivotingrewardfn)$ est le résultat de l'exécution de 4AL
niveau~2 sur le composant $\component$ et la fonction de récompense de
pivotement $\pivotingrewardfn$.

\begin{proposition}\label{pro:level1correct} 
  Soit $LN$ un réseau logique. Supposons que, pour tous les appels à
  4AL niveau~2 effectués par 4AL niveau~1 quand lancé sur $LN$, nous
  avons $V(C,\pivotingrewardfn) = V^*(C,\pivotingrewardfn)$. Alors
  $V(LN) = V^*(LN)$. Si $V(C,\pivotingrewardfn) \leq
  V^*(C,\pivotingrewardfn)$ pour tous les appels à 4AL niveau~2, alors
  $V(LN) \leq V^*(LN)$.
\end{proposition}

Considérons maintenant l'algorithme~\ref{alg:level2}. Notre
description précédente était imprécise en omettant l'argument
additionnel de l'algorithme $\pivotingrewardfn$. Celui-ci s'intègre à
l'algorithme en étant passé, pour chaque sous-réseau sur les chemins
que nous considérons (ligne 7), à l'algorithme~\ref{alg:level3}
qui l'ajoute à la récompense obtenue pour avoir pénétré dans ce
sous-réseau (algorithm~\ref{alg:level3} ligne 4).

$\reward$ agrège les valeurs (lignes 1, 9) sur tous les sous-réseaux
récompensés $N$. Cette agrégation est rendue conservative en enlevant
toutes les récompenses -- les récompenses de pivotement aussi bien que
les récompenses propres des machines individuelles impliquées -- qui
ont déjà été prise en compte (ligne 10). En ce qui concerne les
machines individuelles, l'algorithme~\ref{alg:level2} utilise pour
faire plus court (a) $\rewardfn(N) > 0$ (ligne 2) et (b) $\rewardfn(N)
\leftarrow 0$ (ligne 10); (a) signifie qu'il existe $m \in N$ tel que
$\rewardfn(m) > 0$; (b) signifie que $\rewardfn(m) \leftarrow 0$ pour
tout $m \in N$. En ce qui concerne les récompenses de pivotement,
notons que la ligne 10 de l'algorithme~\ref{alg:level2} modifie la
fonction $\pivotingrewardfn$ maintenue par
l'algorithme~\ref{alg:level1}. Cela ne créé pas de conflits puisque,
quand l'algorithme~\ref{alg:level1} appelle
l'algorithme~\ref{alg:level2} sur le composant $C$, tous les
descendants de $C$ dans $LN$ ont déjà été traités, et donc en
particulier l'algorithme~\ref{alg:level1} ne fera plus de mises à
jour de la valeur de $\pivotingrewardfn(N)$, quelque soit $N \in C$.

Par $\component_\start$ (ligne 4) nous désignons l'ensemble $\{N \in
\component \mid \exists N' \in LN, N' \not \in \component: (N',N) \in
LN\}$ de sous-réseaux qui servent d'entrée dans $\component$ (par
exemple, $N_1$ et $N_3$ pour $C_1$ dans la
figure~\ref{fig:4ALlevels123} (b)). Notons à la ligne 4 que le chemin
$P$ commence avec un pare-feu $F_0$. Pour comprendre cela, observons
la situation considérée. L'algorithme suppose que le parent $N$ de
$\component$ (\start, pour le composant $C_1$ dans la
figure~\ref{fig:4ALlevels123} (b)) est sous contrôle. Mais alors, pour
pénétrer $\component$, nous avons encore besoin de traverser un arc de
$N$ vers $\component$. $F_0$ est le pare-feu sur l'arc choisi par $P$
($F^*_1$ ou $F^*_3$ dans la figure~\ref{fig:4ALlevels123} (b)).

Les appels au niveau~3 (ligne 7) comprennent le réseau $N_i$ dans
lequel pénétrer, le pare-feu $F_{i-1}$ qui doit être traversé pour
cela, la récompense de pivotement de $N_i$, ainsi que la récompense de
chemin actuelle $\reward(P)$ qui est rétro-propagée le long du
chemin. Clairement, c'est équivalent à la séquence d'attaques requise
pour exécuter $P$ et récolter toutes les récompenses de pivotement
associées à une telle attaque. Ainsi, l'agrégation à travers les
sous-réseaux $N$ étant conservative, nous obtenons:

\begin{proposition}\label{pro:level2conservative} 
  Soit $\component$ un composant bi-connecté, et soit
  $\pivotingrewardfn$ une fonction de récompense de
  pivotement. Supposons que, pour tout appel à 4AL niveau~3 effectué
  par 4AL niveau~2 quand exécuté sur $(\component,\pivotingrewardfn)$,
  nous avons $$V(F,N,\pivotingreward,\pathreward) \leq
  V^*(F,N,\pivotingreward,\pathreward).$$
  Alors $$V(C,\pivotingrewardfn) \leq V^*(C,\pivotingrewardfn).$$
\end{proposition}

Les algorithmes~\ref{alg:level3} et~\ref{alg:level4} ne devraient pas
nécessiter d'explications, étant donnée notre discussion
précédente. Notons juste que la récompense de pivotement
$\pivotingreward$ est représentée par l'arc de $m$ vers $C_3$ dans la
figure~\ref{fig:4ALlevels123} (c), laquelle est prise en compte
simplement en l'additionnant à la valeur de $m$
(algorithme~\ref{alg:level3} ligne 4). La récompense de chemin
$\pathreward$ (non illustrée dans la figure~\ref{fig:4ALlevels123}
(c)) est aussi ajoutée à la valeur de $m$ (algorithme~\ref{alg:level3}
ligne 4). Maximiser sur les attaques de machines individuelles $m$
est, de manière évidente, une approximation conservative parce que les
stratégies d'attaque sont libres de choisir $m$. Ainsi:

\begin{proposition}\label{pro:level3conservative} 
  Soit $F$ un pare-feu, $N$ un sous-réseau, $\pivotingreward$ une
  récompense de pivotement, et $\pathreward$ une récompense de
  chemin. Supposons que, pour tous les appels à 4AL niveau~4 effectués
  par 4AL niveau~3 quand exécuté sur
  $(F,N,\pivotingreward,\pathreward)$, nous avons $$V(F,m,\reward)
  \leq V^*(F,m,\reward).$$ Alors $$V(F,N,\pivotingreward,\pathreward)
  \leq V^*(F,N,\pivotingreward,\pathreward).$$
\end{proposition}

\subsection{Construction de la politique}

Au niveau~1, la politique globale est construite à partir de politiques
du niveau~2 simplement en suivant la décomposition arborescente: en
partant de la racine de l'arbre, on exécute les politiques de niveau~2
pour tous les composants atteints (dans n'importe quel ordre); une
fois que le piratage d'un composant a réussi, les composants enfants
respectifs deviennent atteints. Au niveau~2, c'est-à-dire à
l'intérieur d'un composant bi-connecté $C$, la politique correspond à
l'ensemble des chemins $P$ considérés par
l'algorithme~\ref{alg:level2}. Chaque $P$ est traité tour à tour. Pour
chaque nœud $N$ de $P$ (jusqu'à échec à entrer dans ce sous-réseau),
nous appelons la politique de niveau~3 correspondante.

Au niveau~3, c'est-à-dire en considérant un seul sous-réseau $N$,
notre politique attaque simplement la machine $m\in N$ qui donnait le
maximum dans l'algorithme~\ref{alg:level3}. La politique attaque
d'abord $m$ à travers le pare-feu en utilisant la politique de
niveau~4 associée. Dans le cas où l'attaque réussit, la politique
attaque les machines restantes $m'\in N$ dans n'importe quel ordre
(c'est-à-dire, pour chaque $m'$, nous exécutons la politique de
niveau~4 associée jusqu'à terminaison). Au niveau~4, la politique est
une politique POMDP classique retournée par notre résolveur de POMDP.

%% file: Images/4AL1.pdf_t
\begin{picture}(0,0)%
\includegraphics{4AL1.pdf}%
\end{picture}%
\setlength{\unitlength}{4144sp}%
\begingroup\makeatletter\ifx\SetFigFont\undefined%
\gdef\SetFigFont#1#2#3#4#5{%
  \reset@font\fontsize{#1}{#2pt}%
  \fontfamily{#3}\fontseries{#4}\fontshape{#5}%
  \selectfont}%
\fi\endgroup%
\begin{picture}(3526,2491)(2723,-2454)
\put(5431,-1096){\makebox(0,0)[b]{\smash{{\SetFigFont{9}{10.8}{\rmdefault}{\mddefault}{\updefault}{\color[rgb]{0,0,0}$C_5$}%
}}}}
\put(2956,-2311){\makebox(0,0)[b]{\smash{{\SetFigFont{9}{10.8}{\rmdefault}{\mddefault}{\updefault}{\color[rgb]{0,0,0}$C_2$}%
}}}}
\put(3586,-2311){\makebox(0,0)[b]{\smash{{\SetFigFont{9}{10.8}{\rmdefault}{\mddefault}{\updefault}{\color[rgb]{0,0,0}$C_3$}%
}}}}
\put(4531,-2311){\makebox(0,0)[b]{\smash{{\SetFigFont{9}{10.8}{\rmdefault}{\mddefault}{\updefault}{\color[rgb]{0,0,0}$C_4$}%
}}}}
\put(5341,-2311){\makebox(0,0)[b]{\smash{{\SetFigFont{9}{10.8}{\rmdefault}{\mddefault}{\updefault}{\color[rgb]{0,0,0}$C_6$}%
}}}}
\put(6016,-2311){\makebox(0,0)[b]{\smash{{\SetFigFont{9}{10.8}{\rmdefault}{\mddefault}{\updefault}{\color[rgb]{0,0,0}$C_7$}%
}}}}
\put(4501,-286){\makebox(0,0)[b]{\smash{{\SetFigFont{9}{10.8}{\rmdefault}{\mddefault}{\updefault}{\color[rgb]{0,0,0}*}%
}}}}
\put(3406,-1231){\makebox(0,0)[b]{\smash{{\SetFigFont{9}{10.8}{\rmdefault}{\mddefault}{\updefault}{\color[rgb]{0,0,0}$C_1$}%
}}}}
\end{picture}%

%% file: Images/4AL2.pdf_t
\begin{picture}(0,0)%
\includegraphics{4AL2.pdf}%
\end{picture}%
\setlength{\unitlength}{4144sp}%
\begingroup\makeatletter\ifx\SetFigFont\undefined%
\gdef\SetFigFont#1#2#3#4#5{%
  \reset@font\fontsize{#1}{#2pt}%
  \fontfamily{#3}\fontseries{#4}\fontshape{#5}%
  \selectfont}%
\fi\endgroup%
\begin{picture}(3444,2941)(2715,-2319)
\put(4501,299){\makebox(0,0)[b]{\smash{{\SetFigFont{9}{10.8}{\rmdefault}{\mddefault}{\updefault}{\color[rgb]{0,0,0}$*$}%
}}}}
\put(2948,-2176){\makebox(0,0)[b]{\smash{{\SetFigFont{9}{10.8}{\rmdefault}{\mddefault}{\updefault}{\color[rgb]{0,0,0}$C_2$}%
}}}}
\put(5926,-2176){\makebox(0,0)[b]{\smash{{\SetFigFont{9}{10.8}{\rmdefault}{\mddefault}{\updefault}{\color[rgb]{0,0,0}$C_3$}%
}}}}
\put(4501,-1951){\makebox(0,0)[b]{\smash{{\SetFigFont{9}{10.8}{\rmdefault}{\mddefault}{\updefault}{\color[rgb]{0,0,0}$N_2$}%
}}}}
\put(3916,-1096){\makebox(0,0)[b]{\smash{{\SetFigFont{9}{10.8}{\rmdefault}{\mddefault}{\updefault}{\color[rgb]{0,0,0}$N_1$}%
}}}}
\put(5086,-1096){\makebox(0,0)[b]{\smash{{\SetFigFont{9}{10.8}{\rmdefault}{\mddefault}{\updefault}{\color[rgb]{0,0,0}$N_3$}%
}}}}
\put(3331,-646){\makebox(0,0)[b]{\smash{{\SetFigFont{12}{14.4}{\rmdefault}{\mddefault}{\updefault}{\color[rgb]{0,0,0}$C_1$}%
}}}}
\put(4449,-953){\makebox(0,0)[b]{\smash{{\SetFigFont{9}{10.8}{\rmdefault}{\mddefault}{\updefault}{\color[rgb]{0,0,0}$F^1_3$}%
}}}}
\put(4182,-93){\makebox(0,0)[b]{\smash{{\SetFigFont{9}{10.8}{\rmdefault}{\mddefault}{\updefault}{\color[rgb]{0,0,0}$F^*_1$}%
}}}}
\put(4825,-84){\makebox(0,0)[b]{\smash{{\SetFigFont{9}{10.8}{\rmdefault}{\mddefault}{\updefault}{\color[rgb]{0,0,0}$F^*_3$}%
}}}}
\put(3996,-1519){\makebox(0,0)[b]{\smash{{\SetFigFont{9}{10.8}{\rmdefault}{\mddefault}{\updefault}{\color[rgb]{0,0,0}$F^1_2$}%
}}}}
\put(4989,-1539){\makebox(0,0)[b]{\smash{{\SetFigFont{9}{10.8}{\rmdefault}{\mddefault}{\updefault}{\color[rgb]{0,0,0}$F^3_2$}%
}}}}
\end{picture}%

%% file: Images/4AL3.pdf_t
\begin{picture}(0,0)%
\includegraphics{4AL3.pdf}%
\end{picture}%
\setlength{\unitlength}{4144sp}%
\begingroup\makeatletter\ifx\SetFigFont\undefined%
\gdef\SetFigFont#1#2#3#4#5{%
  \reset@font\fontsize{#1}{#2pt}%
  \fontfamily{#3}\fontseries{#4}\fontshape{#5}%
  \selectfont}%
\fi\endgroup%
\begin{picture}(3349,3210)(4640,-1964)
\put(7756,-376){\makebox(0,0)[b]{\smash{{\SetFigFont{12}{14.4}{\rmdefault}{\mddefault}{\updefault}{\color[rgb]{0,0,0}$C_3$}%
}}}}
\put(4853,-386){\makebox(0,0)[b]{\smash{{\SetFigFont{12}{14.4}{\rmdefault}{\mddefault}{\updefault}{\color[rgb]{0,0,0}$N_1$}%
}}}}
\put(6526,524){\makebox(0,0)[b]{\smash{{\SetFigFont{12}{14.4}{\rmdefault}{\mddefault}{\updefault}{\color[rgb]{0,0,0}$m$}%
}}}}
\put(6526,-1456){\makebox(0,0)[b]{\smash{{\SetFigFont{12}{14.4}{\rmdefault}{\mddefault}{\updefault}{\color[rgb]{0,0,0}$m'_k$}%
}}}}
\put(6526,-916){\makebox(0,0)[b]{\smash{{\SetFigFont{12}{14.4}{\rmdefault}{\mddefault}{\updefault}{\color[rgb]{0,0,0}$\dots$}%
}}}}
\put(6526,-376){\makebox(0,0)[b]{\smash{{\SetFigFont{12}{14.4}{\rmdefault}{\mddefault}{\updefault}{\color[rgb]{0,0,0}$m'_1$}%
}}}}
\put(5310,-768){\makebox(0,0)[b]{\smash{{\SetFigFont{12}{14.4}{\rmdefault}{\mddefault}{\updefault}{\color[rgb]{0,0,0}$F^ 1_3$}%
}}}}
\put(5350,  7){\makebox(0,0)[b]{\smash{{\SetFigFont{12}{14.4}{\rmdefault}{\mddefault}{\updefault}{\color[rgb]{0,0,0}$F^ 1_3$}%
}}}}
\put(5588,-399){\makebox(0,0)[b]{\smash{{\SetFigFont{12}{14.4}{\rmdefault}{\mddefault}{\updefault}{\color[rgb]{0,0,0}$F^ 1_3$}%
}}}}
\put(6950,-582){\makebox(0,0)[lb]{\smash{{\SetFigFont{12}{14.4}{\rmdefault}{\mddefault}{\updefault}{\color[rgb]{0,0,0}$\emptyfirewall$}%
}}}}
\put(6683, 72){\makebox(0,0)[lb]{\smash{{\SetFigFont{12}{14.4}{\rmdefault}{\mddefault}{\updefault}{\color[rgb]{0,0,0}$\emptyfirewall$}%
}}}}
\put(7016,1040){\makebox(0,0)[b]{\smash{{\SetFigFont{12}{14.4}{\rmdefault}{\mddefault}{\updefault}{\color[rgb]{0,0,0}$N_3$}%
}}}}
\end{picture}%

%% file: Algorithms/4AL1.tex
\ifthenelse{\isundefined{\jfpda}}{
\begin{minipage}{0.99\columnwidth}
}{
\begin{minipage}{0.49\columnwidth}
}
\vspace{-2.4cm}
\begin{algorithm}[H]
  \caption{Level 1 (Decomposing the Network)}
  \label{alg:level1}

  \KwIn{$LN$: Logical Network.}

  \KwOut{Approximation $V$ of expected value $V^*$ of attacking $LN$
    from controlled machine \start.}

  \acom{Decompose $LN$ into tree $DLN$ of biconnected components,
    rooted at \start; see text for ``clean-up''.} $DLN \gets
  $HopcroftTarjan$(LN)$\; 

  Set tree root to \start\ and \emph{clean-up} $LN$ and $DLN$\;


  $\component_1, \dots, \component_k \gets$ a topological ordering of
  $DLN$\; 

  Intitialize pivoting reward $\pivotingrewardfn(N)$ for all $N \in
  LN$ to $0$\;

   \For{$i = k, ..., 1$}{ 

     \acom{Call Level 2 to attack each component.}  $V(\component_i)
     \gets$Level2$(\component_i,\pivotingrewardfn)$\;

     \acom{Propagate expected reward.} $N \gets$ the parent of
     $\component_i$ in $LN$\; 

     $\pivotingrewardfn(N) \gets \pivotingrewardfn(N) +
     V(\component_i)$\;

   }
   
  \Return{$\pivotingrewardfn(\start)$}
\end{algorithm}
\end{minipage}

%% file: Algorithms/4AL2.tex
\ifthenelse{\isundefined{\jfpda}}{
\begin{minipage}{0.99\columnwidth}
}{
\begin{minipage}{0.49\columnwidth}
}
\begin{algorithm}[H]
  \caption{Level 2 (Attacking Components)}
  \label{alg:level2}

  \KwIn{Biconnected component $\component$, reward function
    $\pivotingrewardfn$.}

  \KwOut{Approximation $V$ of expected value $V^*$ of attacking
    $\component$, given its parent is controlled and its pivoting
    rewards are $\pivotingrewardfn$.}

  
  $\reward \gets 0$\;

  \acom{Account for each rewarded vertex $N$.}

  \While{$ \exists N \in \component$ s.t.\ $\rewardfn(N) > 0$ or
    $\pivotingrewardfn(N)>0$}{

    $P \gets \langle \rangle$; $\reward(P) \gets 0$; $P(N) \gets P$\;

    \acom{Maximize over all simple paths (no repeated vertices) from
      an entry vertex to $N$.}

    \ForEach{simple path $P$ of the form $\xrightarrow{F_0} N_1
      \xrightarrow{F_1} N_2 \dots \xrightarrow{F_{k-1}} N_k = N$ where
      $N_1, \dots, N_k \in \component$ and $N_1 \in
      \component_\start$}{

      \acom{Propagate rewards along $P$, calling Level 3 for attack on
        each subnetwork.}

      $\reward(P) \gets 0$\; 

      \For{$i = k, ..., 1$}{

        $\reward(P) \gets$ Level3$(N_{i},F_{i-1},\pivotingrewardfn(N_{i}),\reward(P))$\; 

      }
      
      $P(N) \gets \text{arg max}(\reward(P(N)),\reward(P))$\;

    }

    $\reward \gets \reward + \reward(P(N))$\;

    $\rewardfn(N_i), \pivotingrewardfn(N_i) \gets 0$ for all vertices $N_i$ on $P(N)$\;

  }

  \Return{$\reward$}

\end{algorithm}
\end{minipage}

%% file: Algorithms/4AL3.tex
\ifthenelse{\isundefined{\jfpda}}{
\begin{minipage}{0.99\columnwidth}
}{
\begin{minipage}{0.49\columnwidth}
}
\vspace{-2.43cm}
\begin{algorithm}[H]
  \caption{Level 3 (Attacking Subnetworks)}
  \label{alg:level3}

  \KwIn{Firewall $F$, subnetwork $N$, rewards $\pivotingreward$,
    $\pathreward$.}

  \KwOut{Approximation $V$ of expected value $V^*$ of attacking $N$
    through $F$, given $F$ is reached, $N$'s pivoting reward is
    $\pivotingreward$, and the path reward behind $N$ is
    $\pathreward$.}

  $\reward \gets 0$\;

  \acom{Maximize over reward obtained when hacking first into a
    particular machine $m \in N$.}

  \ForEach{$m \in N$}{

    $\reward(m) \gets \rewardfn(m)$\;

    \acom{After breaking $m$, we can pivot behind $N$, and reach all
      $m \neq m' \in N$ without $F$.}

    $\reward(m) \gets \reward(m) + \pivotingreward + \pathreward$\;

    \ForEach{$m \neq m' \in N$}{

      $\reward(m) \gets \reward(m) +
      $Level4$(m',\emptyfirewall,\rewardfn(m'))$\;

    }

    $\reward \gets \max(\reward,$ Level4$(m,F,\reward(m)))$\;

  }

  \Return{$\reward$}
\end{algorithm}
\end{minipage}

%% file: Algorithms/4AL4.tex
\ifthenelse{\isundefined{\jfpda}}{
\begin{minipage}{0.99\columnwidth}
}{
\begin{minipage}{0.49\columnwidth}
}
\vspace{0.4cm}
\begin{algorithm}[H]
  \caption{Level 4 (Attacking Individual Machines)}
  \label{alg:level4}

  \KwIn{Firewall $F$, machine $m$, reward $\reward$.}

  \KwOut{Approximation $V$ of expected value $V^*$ of attacking $m$
    through $F$, given $m$ is reached and the current reward of
    breaking it is $\reward$.}

  \If{$(m,F,\reward)$ is cached}{ \Return{$V(m,F,\reward)$} }

  $M \gets$createPOMDP$(m,F,\reward)$\; 

  $V \gets$solvePOMDP$(M)$\;

  Cache $(m,F,\reward)$ with $V$\;

  \Return{$V$}

\end{algorithm}
\end{minipage}

%% file: experiments-fr.tex
\ifthenelse{\isundefined{\jfpda}}{
\begin{figure*}[t]
\begin{tabular}{cc}
\includegraphics[width=\columnwidth]{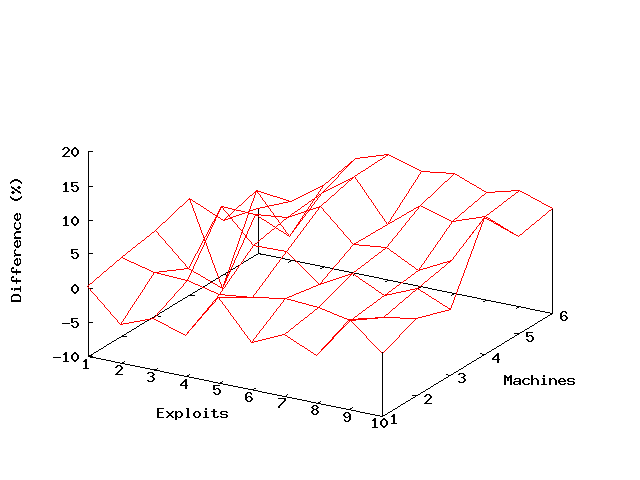} &
\includegraphics[width=\columnwidth]{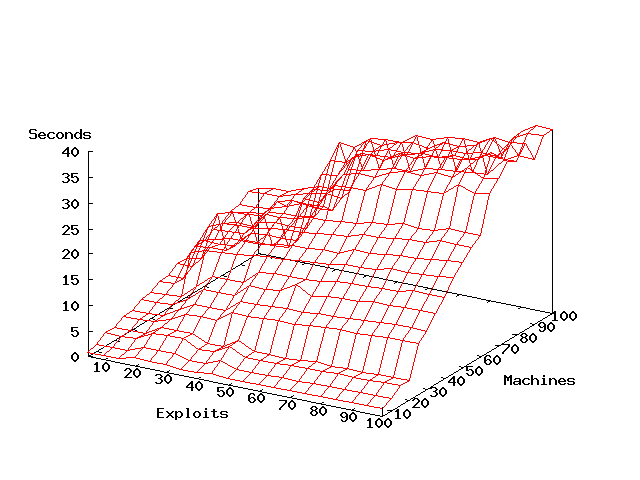}\\[-0.6cm]
(a) Comparaison de la qualité des attaques.  & (b) Temps d'exécution de 4AL. 
\end{tabular}
\vspace{-0.2cm}
\caption{Résultats expérimentaux de 4AL comparé à un modèle POMDP global.}
  \label{fig:experiments}
  \vspace{-0.5cm}
\end{figure*}
}{
  \begin{figure}[t]
    \begin{tabular}{cc}
      \includegraphics[width=0.5\linewidth]{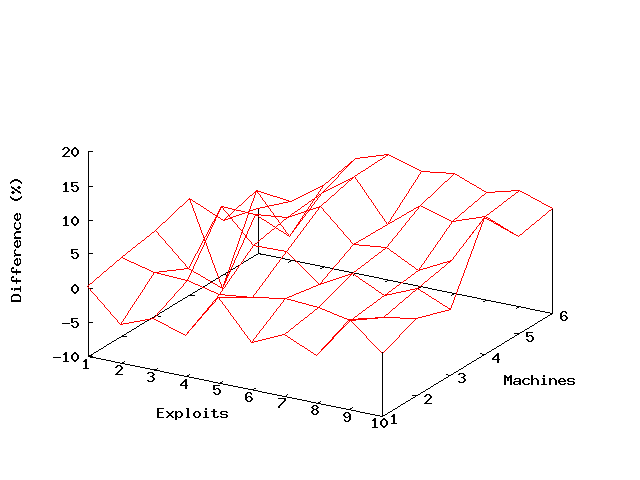} &
      \includegraphics[width=0.5\linewidth]{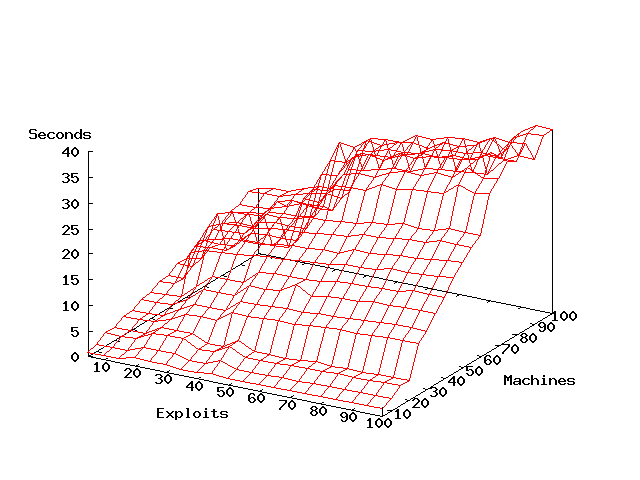}\\[-0.6cm]
      (a) Comparaison de la qualité des attaques.  & (b) Temps d'exécution de 4AL. 
    \end{tabular}
    \caption{Résultats expérimentaux de 4AL comparé à un modèle POMDP global.}
    \label{fig:experiments}
  \end{figure}
}

\section{Expérimentations}
\label{experiments}

Nous avons évalué 4AL contre le modèle POMDP ``global'', encodant le
problème de l'attaque tout entier dans un seul POMDP. Les
expérimentations sont lancées sur une machine avec un CPU Intel Core2
Duo à 2,2~GHz et avec 3~Go de RAM. L'algorithme 4AL est implémenté en
Python. Pour résoudre et évaluer les POMDP générés par le niveau~4,
nous utilisons la boîte à outils APPL.\footnote{APPL 0.93 sur
  http://bigbird.comp.nus.edu.sg/pmwiki/farm/appl/}

\subsection{Scénarios de test}
\label{experiments:test-suite}

\begin{figure}[htb]
\vspace{-0.0cm}
\centering
\includegraphics[width=0.95\columnwidth]{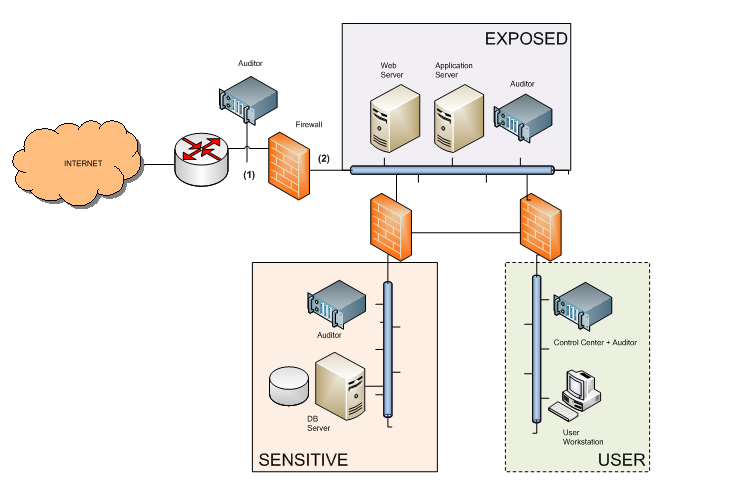}
  \vspace{-0.3cm}
\caption{Structure du réseau dans notre jeu de tests.}
  \label{fig:scenario}
  \vspace{-0.6cm}
\end{figure}

Notre scénario de test repose sur la structure de réseau montrée dans
la figure~\ref{fig:scenario}. L'attaque commence depuis Internet
(\start\ est le nuage dans le coin en haut à gauche). Le réseau
consiste en 3 zones -- \emph{Exposed, Sensitive, User} -- séparées par
des pare-feux. En interne, Exposed comme Sensitive sont complètement
connectées (c'est-à-dire que ces zones sont des sous-réseaux), alors
que User consiste en un arbre de sous-réseaux séparés par des
pare-feux vides. Seules deux machines sont récompensées, une dans
Sensitive (récompense 9000) et une dans un sous-réseau feuille de User
(récompense 5000). Le coût des scans de ports et des exploits est de
10, le coût des détections d'OS de 50. Nous permettons de faire
croître le nombre de machines $|M|$ en distribuant, toutes les 40
machines, la première dans Exposed, la seconde dans Sensitive, et les
38 restantes dans User. Les exploits sont pris dans la base de données
de Core Security. Le nombre d'exploits $|E|$ est passé à l'échelle en
distribuant ceux-ci sur 13 modèles, et en affectant à chaque machine
$m$ un tel modèle en tant que $I(m)$ (la configuration à l'instant du
dernier pentest). La croyance initiale $b_0(I,T)$, où $T$ est le temps
passé depuis le dernier pentest, est alors généré tel que décrit.

Les paramètres fixés ici (récompenses, coûts d'action, distribution
des machines sur les zones, nombre de modèles) sont estimés à partir
de l'expérience pratique de Core Security. La structure du réseau et
les exploits sont réalistes, et sont utilisés pour des tests
industriels dans cette compagnie. La principale faiblesse du scénario
réside dans l'approximation des mises à jour logicielles soustendant
$b_0(I,T)$. Dans l'ensemble, le scénario est encore simplifié, mais
il est naturel et approche la complexité des tests de pénétration réels.

Par manque d'espace, dans ce qui suit nous faisons passer à l'échelle
seulement $|M|$ et $|E|$, en fixant $T = 50$. Cette dernière valeur
est difficile: le pentesting est typiquement accomplit à peu près une
fois par mois; de plus petits $T$ sont plus faciles à résoudre
puisqu'il y a moins d'incertitude.

\subsection{Perte en approximation}
\label{experiments:approximation}

La figure~\ref{fig:experiments} (a) montre la perte relative de
qualité quand on exécute 4AL au lieu d'une solution POMDP globale,
pour des valeurs de $|E|$ et $|M|$ où cette dernière est faisable.
Nous montrons
$\mathit{quality}(\mathit{global\mbox{-}POMDP})-\mathit{quality}(\mathit{4AL})$
en pourcent de $\mathit{quality}(\mathit{global\mbox{-}POMDP})$. La
qualité de la politique est ici estimée en exécutant 2000
simulations. Cette mesure subit une variance, laquelle est presque
plus forte que le très petit avantage en qualité de la solution POMDP
globale. La perte maximale pour n'importe quelle combinaison de $|E|$
et $|M|$ est de $14,1\%$ (pour $|E|=7$, $|M|=6$), la perte moyenne sur
toutes les combinaisons est de $1,96\%$. La perte moyenne croît de
façon monotone selon $|M|$, de $-1,14\%$ pour $|M|=1$ à $4,37\%$ for
$|M|=6$. Selon $|E|$, le comportement est moins régulier; la perte
moyenne maximale, $5,4\%$, est obtenue en fixant $|E|=5$.

\subsection{Passage à l'échelle}
\label{experiments:scaling}

La figure~\ref{fig:experiments} (b) montre le temps d'éxecution de 4AL
quand on passe à de plus grandes valeurs de $|E|$ et $|M|$. Le
comportement de passage à l'échelle selon $|M|$ reflète clairement le
fait que 4AL est polynomial en ce paramètre, sauf pour la taille des
composants bi-connectés (ici $4$). Faire croître $E$ donne des POMDP
mono-machines plus difficiles, résultant parfois en une croissance
forte du temps de calcul. Pourtant, même avec $|M|$ et $|E|$ tous deux
proches de $100$, ce qui est une taille réaliste en pratique, le temps
d'exécution est toujours sous les $37$ secondes.

%% file: conclusion-fr.tex
\section{Conclusion}
\label{conclusion}

Nous avons conçu un modèle POMDP pour les tests de pénétration qui
permet de représenter naturellement beaucoup des caractéristiques
de cette application, en particulier la connaissance incomplète de la
configuration du réseau, ainsi que les dépendences entre les
différentes attaques possibles et les pare-feux. A la différence des
méthodes précédentes, cette approche est capable de mélanger
intelligemment scans et exploits. Si cette solution exacte ne passe
pas à l'échelle, de grands réseaux peuvent être traités par un
algorithme de décomposition. Nos résultats expérimentaux suggèrent que
c'est accompli pour une faible perte en qualité par rapport à une
solution POMDP globale.

Une importante question ouverte est dans quelle mesure notre approche
POMDP+decomposition est plus efficace en termes de coûts que la
solution par planification classique actuellement employée par Core
Security. Notre prochaine étape sera de répondre à cette question
expérimentalement, en comparant la qualité des attaques de 4AL avec
celle de la politique qui exécute des scans extensifs et exécute
ensuite les plans de FF associés à la configuration la plus probable.

4AL est un algorithme spécifique à un domaine et, en tant que tel, ne
contribue pas à la résolution de POMDP en général. A un haut niveau
d'abstraction, son idée peut être comprise comme imposant un schéma
sur la politique construite, restreignant ainsi l'espace des
politiques explorées (et employant des algorithmes dédiés dans chaque
partie du schéma). En cela, cette approche est assez similaire aux
approches de décomposition de POMDP connues (par exemple,
\cite{PinGorThr-uai03,mueller:biundo:ki-11}). Il reste à voir si cette
connexion peut s'avérer fructueuse soit pour de travaux futurs en
planification d'attaques, soit pour la résolution de POMDP de manière
plus générale.

Les principales directions pour des travaux futurs sont de concevoir
des modèles plus précis des mises à jour logicielles (donc obtenant
des calculs plus réalistes de la croyance initiale); d'adapter les
résolveurs de POMDP à ce type particulier de problème, qui a certaines
caractéristiques spécifiques, en particulier l'absence d'actions
non-déterministes et le fait que certaines parties de l'état (par
exemple les systèmes d'exploitation) sont statiques; et de faire
progresser l'application industrielle de cette technologie. Nous
espérons que tout cela inspirera aussi d'autres recherches.

%% file: JFPDA-article.bbl
\begin{thebibliography}{~~~}

\bibitem[\protect\citename{Arce \& McGraw, }2004]{ArcGra04}
{\sc Arce I. \& McGraw G.} (2004).
\newblock Why attacking systems is a good idea.
\newblock {\em IEEE Computer Society - Security \& Privacy Magazine}, {\bf
  2}(4).

\bibitem[\protect\citename{Bertsekas \& Tsitsiklis, }1996]{BerTsi96}
{\sc Bertsekas D. \& Tsitsiklis J.} (1996).
\newblock {\em Neurodynamic Programming}.
\newblock Athena Scientific.

\bibitem[\protect\citename{Boddy {\em et~al.}, }2005]{BodGohHaiHar05}
{\sc Boddy M.~S., Gohde J., Haigh T. \& Harp S.~A.} (2005).
\newblock Course of action generation for cyber security using classical
  planning.
\newblock In {\em Proc. of ICAPS'05}.

\bibitem[\protect\citename{Hansen \& Feng, }2000]{HanFen00}
{\sc Hansen E. \& Feng Z.} (2000).
\newblock Dynamic programming for {POMDPs} using a factored state
  representation.
\newblock In {\em Proceedings of the International Conference on AI Planning
  and Scheduling (AIPS'00)}.

\bibitem[\protect\citename{Hoffmann, }2003]{hoffmann:jair-03}
{\sc Hoffmann J.} (2003).
\newblock The {M}etric-{FF} planning system: Translating ``ignoring delete
  lists'' to numeric state variables.
\newblock {\em Journal of Artificial Intelligence Research}, {\bf 20},
  291--341.

\bibitem[\protect\citename{Hopcroft \& Tarjan, }1973]{HopTar-cACM73}
{\sc Hopcroft J. \& Tarjan R.} (1973).
\newblock Algorithm 447: efficient algorithms for graph manipulation.
\newblock {\em Communications of the ACM}, {\bf 16}, 372--378.

\bibitem[\protect\citename{Kaelbling {\em et~al.}, }1998]{KaeLitCas-aij98}
{\sc Kaelbling L., Littman M. \& Cassandra A.} (1998).
\newblock Planning and acting in partially observable stochastic domains.
\newblock {\em Artificial Intelligence}, {\bf 101}(1--2), 99--134.

\bibitem[\protect\citename{Kurniawati {\em et~al.}, }2008]{KurHsuLee08}
{\sc Kurniawati H., Hsu D. \& Lee W.} (2008).
\newblock {SARSOP}: Efficient point-based {POMDP} planning by approximating
  optimally reachable belief spaces.
\newblock In {\em Robotics: Science and Systems {IV}}.

\bibitem[\protect\citename{Lucangeli {\em et~al.}, }2010]{LucSarRic10}
{\sc Lucangeli J., Sarraute C. \& Richarte G.} (2010).
\newblock Attack planning in the real world.
\newblock In {\em Workshop on Intelligent Security (SecArt 2010)}.

\bibitem[\protect\citename{Monahan, }1982]{Monahan82}
{\sc Monahan G.} (1982).
\newblock A survey of partially observable {Markov} decision processes.
\newblock {\em Management Science}, {\bf 28}, 1--16.

\bibitem[\protect\citename{M{\"u}ller \& Biundo, }2011]{mueller:biundo:ki-11}
{\sc M{\"u}ller F. \& Biundo S.} (2011).
\newblock {HTN}-style planning in relational {POMDPs} using first-order {FSCs}.
\newblock In {\em Proceedings of the 34th German Conference on AI (KI'11)}, p.\
  216--227.

\bibitem[\protect\citename{Pineau {\em et~al.}, }2003]{PinGorThr-uai03}
{\sc Pineau J., Gordon G. \& Thrun S.} (2003).
\newblock Policy-contingent abstraction for robust robot control.
\newblock In {\em Proceedings of the 19th Conference on Uncertainty in
  Articifical Intelligence (UAI'03)}, p.\ 477--484.

\bibitem[\protect\citename{Sarraute {\em et~al.}, }2011a]{SarBufHof11}
{\sc Sarraute C., Buffet O. \& Hoffmann J.} (2011a).
\newblock Penetration testing == {POMDP} solving?
\newblock In {\em Proceedings of the 3rd Workshop on Intelligent Security
  (SecArt'11)}.

\bibitem[\protect\citename{Sarraute {\em et~al.}, }2011b]{SarRicLuc11}
{\sc Sarraute C., Richarte G. \& Lucangeli J.} (2011b).
\newblock An algorithm to find optimal attack paths in nondeterministic
  scenarios.
\newblock In {\em ACM Workshop on Artificial Intelligence and Security
  (AISec'11)}.

\end{thebibliography}
